\documentclass[runningheads]{llncs}
\usepackage[T1]{fontenc}
\usepackage{graphicx}
\usepackage{tabularx}
\begin{document}
%
\title{Prompt me a Dataset: An investigation of text-image prompting for historical image dataset creation using foundation models}
\titlerunning{Prompt me a Dataset}

\author{Hassan El-Hajj\inst{1,2}\orcidID{0000-0001-6931-7709} \and
Matteo Valleriani\inst{1,2,3,4}\orcidID{0000-0002-0406-7777}}
\authorrunning{El-Hajj. H., and Valleriani, M.}
\institute{Max Planck Institute for the History of Science, Boltzmannstr. 22, Berlin, 14195, Germany \and BIFOLD -- Berlin Institute for the Foundations of Learning and Data, Berlin, 10587, Germany
\and The Cohn Institute for the History and Philosophy of Science and Ideas, Faculty of Humanities, Tel-Aviv University, Tel-Aviv, 6997801, Israel 
\and Institute of History and Philosophy of Science, Technology, and Literature, Faculty I, Technische Universität Berlin, Straße des 17. Juni 135, Berlin, 10623, Germany\\
\email{\{hhajj,valleriani\}@mpiwg-berlin.mpg.de}}

%

%
\maketitle              
\begin{abstract}
In this paper, we present a pipeline for image extraction from historical documents using foundation models, and evaluate  text-image prompts and their effectiveness on humanities datasets of varying levels of complexity. The motivation for this approach stems from the high interest of historians in visual elements printed alongside historical texts on the one hand, and from the relative lack of well-annotated datasets within the humanities when compared to other domains. We propose a sequential approach that relies on GroundDINO and Meta's Segment-Anything-Model (SAM) to retrieve a significant portion of visual data from historical documents that can then be used for downstream development tasks and dataset creation, as well as evaluate the effect of different linguistic prompts on the resulting detections.

\end{abstract}
\section{Introduction}\label{text:intro}
Technological advancements of the last decades have led to major digitization efforts focused on historical documents, such as the Google Book Search (GBS) and Open Content Alliance (OCA) \cite{Jones2011}. This rapid growth of digitized historical documents has paved the way for computational historical document analysis, allowing researchers to comb through large number of documents and test hypotheses at scale.\par

With the advent of neural networks, new methods of image analysis and information extraction came to light. However, these methods are data intensive, and require a large amount of annotated and curated datasets in order to be trained. The lacuna of such datasets pushed the digital humanities community towards collecting and publishing annotated and curated datasets to facilitate the training of state-of-the-art models. However, given the heterogeneous nature of historical data, and the high degree of inter --and intra-- domain variability, such datasets often cover very specific historical topics and domains, with limited generalization possibilities.\par

In this paper, we propose a pipeline for information extraction from historical documents using image foundation models to support the work of historians. We discuss the current state of research for information extraction pipelines within the humanities in Section \ref{text:SOTA}. In Section \ref{text:Method}, we discuss our current pipeline as well as the current experience within the Max Planck Institute for the History of Science, evaluate it on three datasets in Section \ref{text:PrelEval}, and conclude with an overview of possible extensions to the proposed pipeline in Section \ref{text:FutureDev}.

\section{Current State of the Research}\label{text:SOTA}
Information Extraction (IE), including image (e.g., visual elements) extraction, from historical documents is playing an increasingly important role in formulating historical hypotheses \cite{Valleriani2022Histories}, allowing researchers to tap into a large pool of information that would have been impossible to assemble without computational methods. Meanwhile, there have been many advances in text processing with regard to both printed and handwritten sources \cite{Gaur2015,Ahmad2022,Fischer2020}. In this paper, we tackle the well-addressed technical problem of image extraction from historical documents while relying on foundation models and text prompts. \par

Current approaches to extract images from historical texts can be divided into two main groups: Segmentation and Object Detection approaches. Segmentation approaches often rely on FCN architectures such as U-Net \cite{UNet} or Mask-RCNN \cite{He2017} to generate masks of the desired image region. One of these approaches is the one proposed by \cite{monnier2020docExtractor} to extract images from a wide range of historical texts using a modified U-Net. Another similar approach is proposed by \cite{Ahmad2022} to segment text lines in handwritten historical documents. Numerous further approaches have treated information extraction and, more specifically, image extraction as an object detection problem and tackled it with models such as EfficientDet \cite{Tan2022}, for instance in \cite{Dutta2021}, where the authors extracted images from a corpus of Scottish Chapbooks. \cite{Buettner2022} used instead YOLOv5 \cite{YOLO} to extract different classes of visual elements from a large corpus of early modern books. \par

While the above-mentioned approaches are far from representing a comprehensive review of the current state of image extraction from historical documents, they highlight general trends within the community. Despite their differences, these approaches share a fundamental feature, namely that they were all trained on carefully annotated datasets. This is notable because, in contrast to other industry domains, \textit{annotated} data within the humanities remains relatively scarce due to numerous reason including a lack of expertise (compared to the difficulty of defining classes within heterogeneous data, as well as ambiguous data interpretations to name a few). Many of the approaches discussed above provide their own datasets, such as the Synthetic \textit{SynDoc} dataset presented in \cite{monnier2020docExtractor}, the Chapbook dataset presented by \cite{Dutta2021}, as well as the S-VED \cite{SVED} presented by \cite{Buettner2022,SVED}. The amount of humanities and historical document datasets is continuously growing with numerous datasets covering different aspects of these fields \cite{Nikolaidou2022}; this is also manifested by the growing number of datasets published on Hugging-Face's BigLAM: Big-Science, Libraries, Archives and Museums group \cite{GLAM}.\par

\begin{figure}[h]
    \centering
    \includegraphics[width=\linewidth]{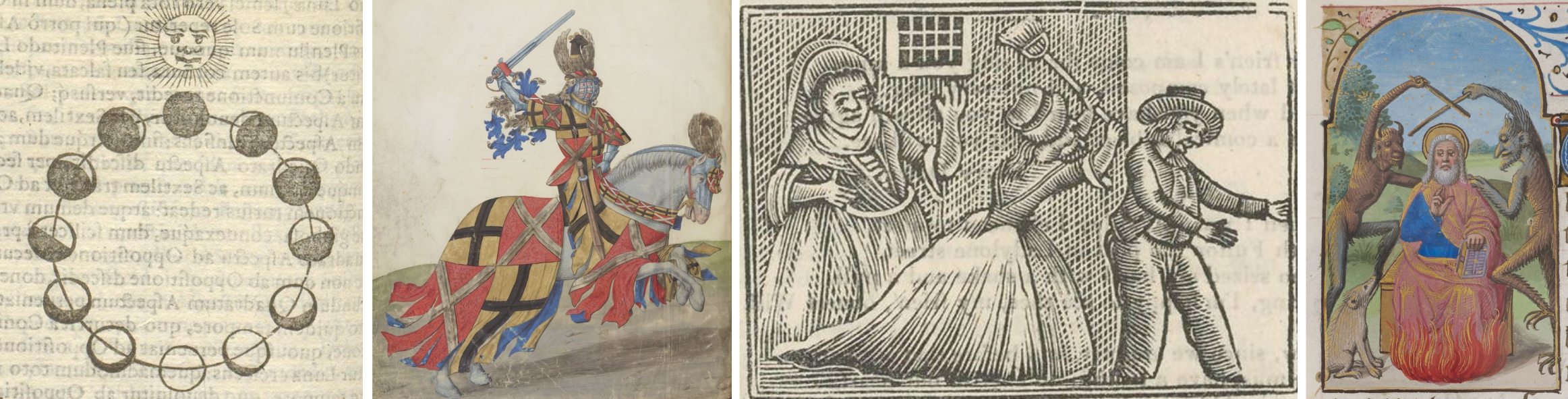}
    \caption{Images of diverse types and styles. (Left to Right) A diagram from the S-VED dataset \cite{Buettner2022}. A colored image from the IlluHistDoc dataset \cite{monnier2020docExtractor}. An image from the Chapbook dataset \cite{Dutta2021}. A miniature from the HORAE dataset \cite{Boillet2019}}
    \label{fig:im_example}
\end{figure}

Despite the consistently growing number of datasets, the high level of heterogeneity of historical documents means that many of these datasets cover a small, often \textit{niche-like}, group of target documents (e.g., Figure \ref{fig:im_example}). This essentially means that the image extraction models -- whether segmentation or object detection -- often perform very well on in-domain data, but suffer from high performance degradation on out-of-domain data. This is clearly shown in the results presented in \cite{Buettner2022}, where the performance of the YOLOv5 model trained on S-VED \cite{SVED} (a dataset containing visual elements from early modern books on astronomy) degrades on out-of-domain datasets, such Mandragore \cite{MandragoreDataset}, a dataset consisting of diverse manuscripts from the Bibliothèque National de France (BnF), or RASM \cite{RASM2019}, a dataset of historical Arabic manuscripts. \par

While these target-specific models are dependent on the presence of well-curated domain datasets, new foundation models are being developed, trained on immensely large datasets and able to perform \textbf{zero-shot} inference, which means that these models are able to perform well on out-of-domain images without requiring extra training.


\section{Pipeline}\label{text:Method}
While stand-alone models perform excellently on in-domain data, we aim to leverage \textit{image} foundation models to help humanities researchers extract visual elements from their datasets as an end goal and, more importantly, quickly generate image datasets from broad data sources without requiring in-domain training. \par

\begin{figure}[h!]
    \centering
    \includegraphics[width=1\linewidth]{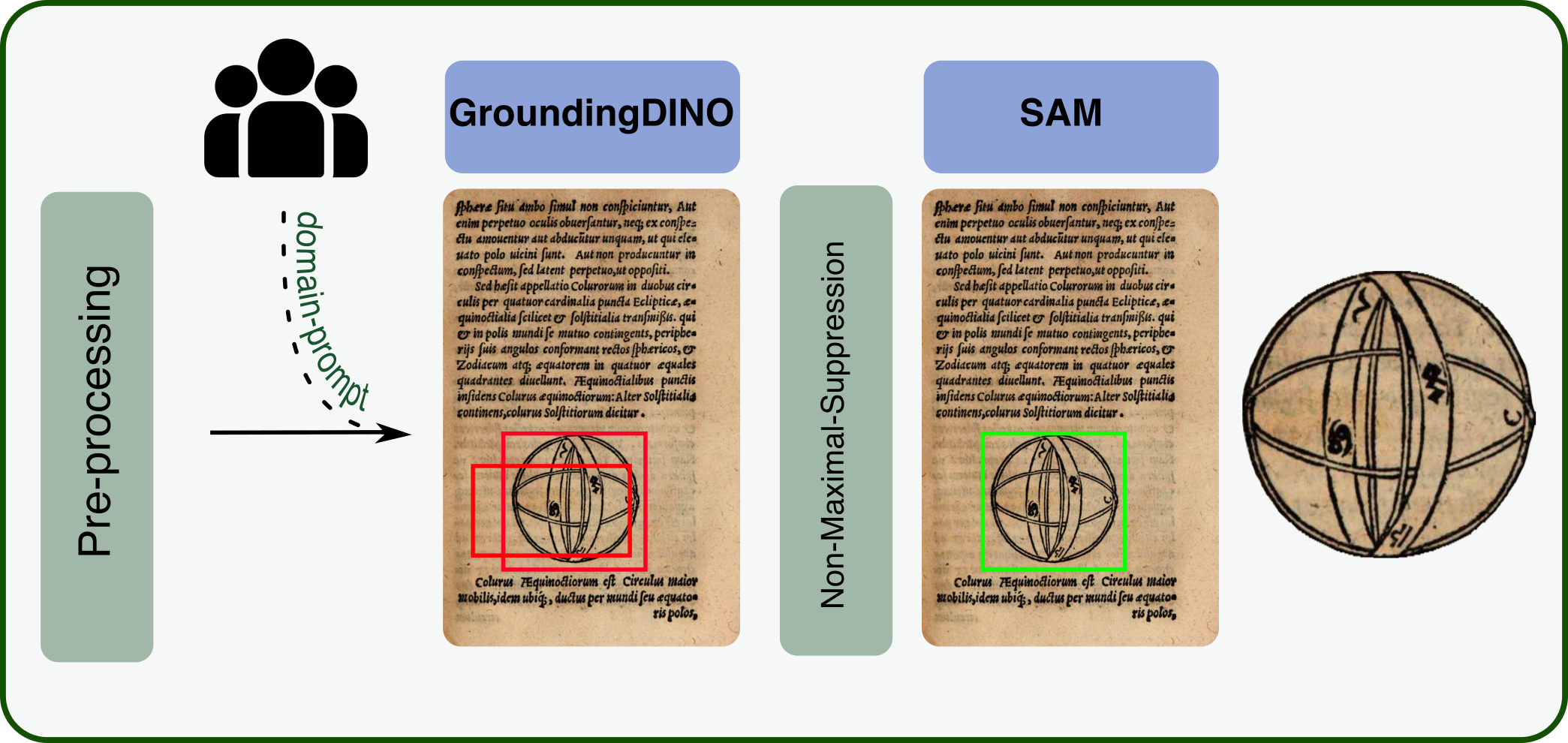}
    \caption{Workflow from out-of-domain data entry on the left towards data extraction as bounding box with GroundingDINO \cite{zhang2022dino} and as masks with SAM \cite{kirillov2023}}
    \label{fig:pipeline}
\end{figure}

This pipeline relies heavily on prompting these foundation models to achieve the best possible image extraction results without the use of domain-specific data. In this case, we chain a GroundingDINO \cite{liu2023} model with a Segment-Anything-Model (SAM) \cite{kirillov2023} to create a pipeline to generate visual element region masks from historical manuscripts (see Figure \ref{fig:pipeline}). \par

GroundingDINO is a model that relies on a Transformer-based end-to-end object detection DINO (DETR with Improved DeNoising Anchor Boxes for End-to-End Object Detection) \cite{zhang2022dino}, and fuses it with a Text-Encoder in order to detect objects based on human language input on open-domain data, achieving good zero-shot results on the COCO dataset \cite{COCO}. 
GroundingDINO takes an image-text input pair, and returns a bounding box that corresponds to the image region that in turns semantically corresponds to its textual counterpart. 
These bounding boxes are then passed on as data prompts to SAM \cite{kirillov2023} in order to segment the desired semantically relevant object.\par

One obvious downside of such models is that despite the fact that they are trained on very large datasets (e.g., SA-1B dataset released by Meta contains 11 million images with 1 billion segmentation masks \cite{kirillov2023}), they often overlook humanities or historically oriented data, excluding data classes needed for manuscript and historical document information extraction. One of the major causes of the current status concerning data is the relatively low number of annotated manuscript and historical text data, as well as the difficulty in retrieving the domain knowledge required to annotate these images (e.g., think of the difference between different classes of images within the same manuscript or printed book). This situation makes it difficult to accommodate or create such data using Mechanical Turk\footnote{Mechanical Turk is an Amazon based marketplace platform where organizations can hire workers, often for relatively low wage, to conduct some low-level work. This service is often used to annotate images and create large datasetes.} workers with little to no historical domain knowledge. To circumvent these shortcomings, we propose utilizing targeted domain-aware prompts that can hone in on the desired objects, and fine-tuning GroundingDINO as part of future developments, as discussed in Section \ref{text:FutureDev}. \par

The proposed pipeline is composed of three blocks: A pre-processing block, an object detection block relying on Prompt engineering GroundingDINO, and a finer segmentation block relying on SAM. The pre-processing block resizes each image to a standard size of 1000x1000 px, and includes an autocontrast step with a 2nd and 98th percentile cutoff. These images are then passed on to the GroundingDINO module with engineered prompts. These prompts are designed in a way to inject domain knowledge  while remaining general enough so that the Feature Enhancer block of the GroundingDINO model is able to fuse text and image features in an efficient way to return reliable results. Examples of these engineered prompts are shown in Section \ref{text:PrelEval}.\par

With the multiple prompt classes, multiple bounding box detections are expected. We thus add a Non-Maximal Suppression module that operates on the selected prompt group classes to ensure that each object is detected once. The cleaned results, i.e., the bounding boxes, are then passed on as box-prompts to the SAM block to return clean segmentation masks of the desired regions. \par

We acknowledge that this pipeline relies on two very large models and might not be efficient to run in production. However, we believe that this approach can drastically increase the amount of data at the disposal of humanities researchers, and allows them to create large datasets using language prompts. For production scenarios or domain-specific requirements, the proposed approach can be used for an initial data collection phase in preparation for the training of bespoke object detector or segmentation models. 


\section{Text-Image Prompt Evaluation}\label{text:PrelEval}

We conduct a preliminary evaluation of the pipeline above on subsets of the S-VED \cite{SVED}, Chapbook \cite{Dutta2021}, and the HORAE datasets \cite{Boillet2019} and report the preliminary results below. These datasets are object detection datasets in historical documents dating from the 15--17th, 17--19th, and the 14th--16th centuries respectively. The S-VED dataset contains four semantically different classes, the most abundant being Content Illustrations which covers visual elements within the body of the text and intend to enrich it. Other classes include Initials, which represent often decorated letters (or drop cap) at the beginning of chapters and paragraphs, Decorations which represents small decorative elements on pages, and Printer's Marks, which represent the emblem of the printer(s) who produced the book in question \cite{Buettner2022}. The Chapbook dataset consists of a single image class representing every image within a text page \cite{Dutta2021} while the pages of the HORAE dataset have the most detailed annotation scheme \cite{Boillet2019}. These cover Miniatures, which are illustrations embedded in the text, Decorations which are elements often placed along the page borders, as well as different types of Initials, such as simple initials (initials differing from the main body of the text in ink and size), decorated initials (initials with purely ornamental decoration style), and historiated initials (initials whose decoration depicts an iconographic element such as a scene or a character) \cite{Boillet2019}.\par 

Beyond the difference in classes, these three datasets represent different types of content which contain different styles of visual elements. The S-VED dataset derives from the Sphere Corpus\footnote{\url{https://sphaera.mpiwg-berlin.mpg.de/}}, which contains scientific books on geocentric astronomy used in pedagogical settings; the Chapbooks were booklets containing popular content ranging from literature, poems, religious texts, and riddles; and the HORAE dataset contains pages from the books of hours, which were a type of handwritten prayer book owned and widely circulated in the late middle ages. The difference between these types of primary sources is naturally reflected in the types of images they contain, with the S-VED containing a large number of orbital diagrams and geometric drawings, the Chapbooks dataset containing a wide range of daily life drawings featuring humans, animals, and in some cases abstract and stylized figures, and the HORAE containing a large amount of decorative elements places around the textual area of the page, as well as a lot of religious illustrations (Figure \ref{fig:im_example}).\par

In our attempt to evaluate the pipeline on the two models, we set the text and image thresholds to 0.35 within the GroundingDINO parameters and perform non-maximal suppression on the output boxes. We also cast all classes of the S-VED into a single visual element class in order to obtain a comparable result between the three chosen datasets. We evaluate the Average Precision (AP) \cite{Cartucho2018} of different language prompts in order to examine their effect on the model's ability to extract the needed information on such as out-of-domain data. \par 

The first language prompt that we applied uses simple language prompts (i.e., single words) to try to extract the visual elements from both datasets. In this case, the prompt is constituted of the single word \textbf{\{figure\}}, which resulted in very good semantically meaningful results on the S-VED dataset, but appears to perform poorly on both the Chapbook and the HORAE dataset (see Table \ref{table:mAP_1000}). \par

\begin{figure}[h!]
    \centering
    \includegraphics[width = \linewidth]{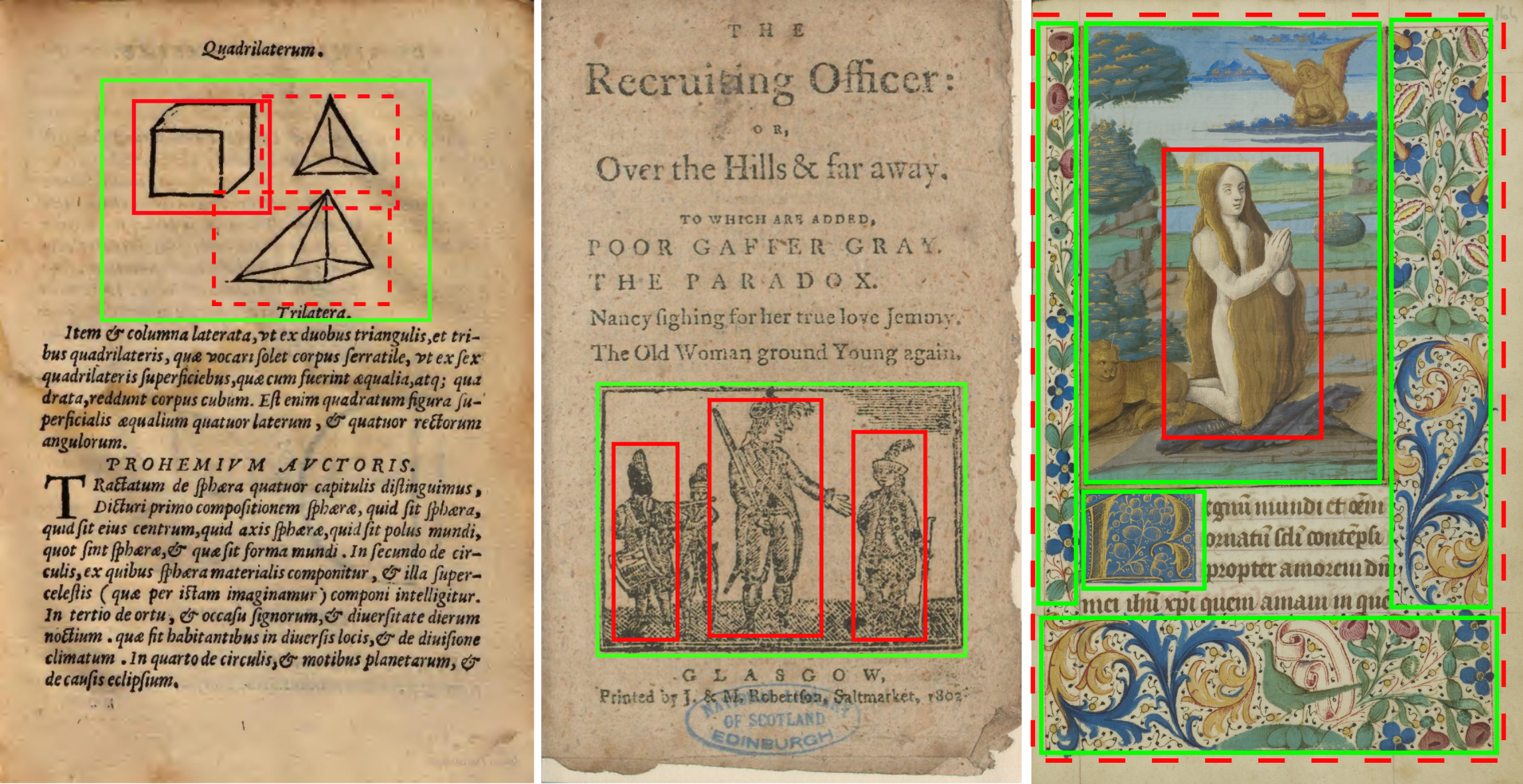}
    \caption{(Left) A page from the S-VED dataset showing multiple detected regions. The solid red line represents the results obtained with the prompt \textbf{\{figure\}}, the dashed red lines represent the two extra boxes detected with the prompt \textbf{\{figure - diagram - geometry - sketch\}}, while the green box represent the ground truth data. (Center) A Chapbook page with multiple region predictions from a \textbf{\{figure\}} prompt in red, and a region prediction given a prompt \textbf{\{image - square - rectangle - photo\}}, in green corresponding to the ground truth. (Right) A HORAE page with ground truth bounding boxes in green showing a miniature, initial, and three decoration boxes around the page. The solid red box shows the prediction from a \textbf{\{figure\}} prompts, while the dashed red line shows the prediction from a \textbf{\{floral - rectangle - flower - decorative - abstract\}} prompt.}
    \label{fig:example}
\end{figure}

We investigate the detection and segmentation results from our pipeline in an effort to improve our prompts and retrieve a larger amount of visual elements. In the S-VED, the error sources were manifold. The first consists of missing small visual elements placed in the marginalia; the second concerns missing abstract geometric shapes that the model did not deem to be a fit for the given textual prompts. However, the highest contributor to the relatively modest AP score reported in Table \ref{table:mAP_1000} is the difference between the bounding boxes that our pipeline considered to be representing a figure, and the bounding boxes created by the annotators of the S-VED, which is highly abstract. A simple example is the presence of three different drawings on an S-VED representing semantically related topics, and thus annotated as a single image by the S-VED annotators. However, relying solely on the image and the text prompt, our pipeline returns multiple smaller bounding boxes with low Intersection-over-Union scores, leading to False Positive results (see Figure \ref{fig:example}).\par

In the case of the Chapbook and HORAE datasets, the main cause of this error was a semantic mismatch between our prompt and the desired outcome. The suggested prompt of \textbf{\{figure\}} has led the GroundingDINO module to return bounding boxes of human \textit{figures} within the visual elements, instead of the desired output of a figure in the literal sense (see Figure \ref{fig:example}). Thus the low score was the result of the identification and segmentation of parts of the visual elements showing a human figure, diverging from the annotated ground truth data. This simple example proved interesting and highlights the multifaceted meaning that a single word prompt could have, and how it could affect the results (see Figure \ref{fig:example}). \par

To inject more domain-knowledge into our prompts, we provide dataset-tailored prompts. For S-VED, we provide a textual prompt that better describes the content of the majority of its visual elements: \textbf{\{figure - diagram - geometry - sketch\}}. In the Chapbook dataset, we focus on identifying the complete visual element, which is often square or rectangular in shape, thus the prompt in this case is \textbf{\{image - square - rectangle - photo\}}, while in the HORAE dataset, where we aim to detect miniatures, we provide a prompt that aims to describe their content based on an our observations, which in this case is \textbf{\{figure - lanscape - scene - square\}}. We re-evaluate the pipeline with the aforementioned linguistic prompts, and notice a small increase in performance for the S-VED, largely due to the detection of some previously missed geometric shapes, and a large increase in performance on the Chapbook and HORAE datasets due to the fact that the new textual prompt aligns with the ground truth annotation scheme.\par

In order to better probe the limits of text-image pairings on a very specific dataset such as historical documents, we attempt to differentiate between the different classes of the S-VED and HORAE dataset. In the case of the S-VED, we focus our attention on differentiating between the Content Illustration and Initial classes, the most abundant classes in the S-VED. In the HORAE dataset we focus on differentiating between the Initial, Decoration, and Miniature classes. In the first case, We prompt the following in order to retrieve the S-VED Initials class \textbf{\{dropcap - decorated letter - large letter\}} and the following for the Content Illustration class \textbf{\{figure - diagram - circle - planets\}}. To differentiate between the HORAE classes, we utilize the same S-VED prompt for the Initials class, and use the following  \textbf{\{floral - rectangle - flower - decorative - abstract\}} and \textbf{\{scene - landscape - square\}} for the Decoration and Miniatures classes respectively. However in the above cases, we see that we have possibly reached the limit of the pre-trained model's text-image understanding, which is likely hindered by our efforts to differentiate the classes using very generalized terms. In the S-VED case, this is noticeable for example when an Initial such as ``O'' or ``D'' is classified as Content Illustration with high confidence due to its circular characteristic. In the HORAE examples, we encountered the same issues with the Initials class; but also faced some problems detecting the decorative elements according to the annotation scheme which divides the decorative elements according to their orientation in the manuscript pages. This meant that while our pipeline often recognizes decorative elements in the page, the detection box does not recognize the distinct decorative elements as per the annotation scheme, resulting in poor performance (see Figure \ref{fig:example} for a clear comparison between the annotation scheme and the detected areas). Such mishaps ultimately resulted in almost random class detections (AP scores of 0.1, 0.08, and 0.12 for the S-VED Initials, HORAE Initials, and Decorative elements respectively), and proved to be an inefficient avenue.\par

\begin{table}[h]
    \centering
    \begin{tabularx}{\columnwidth}{|l|X|r|}
    \hline
         \textbf{Dataset} & \textbf{Prompt} & \textbf{AP} \\
         \hline
         S-VED & \{figure\} & 0.42\\
         S-VED & \{figure - diagram - geometry - sketch\} & \textbf{0.51}\\
         \hline
         Chapbook & \{figure\} & 0.19\\
         Chapbook & \{image - square - rectangle - photo\} & \textbf{0.82}\\
         \hline
         HORAE & \{figure\} & 0.15\\
         HORAE & \{figure - lanscape - scene - square \} & \textbf{0.74}\\
         \hline
    \end{tabularx}
    \caption{Average Precision score for object detection on a subset of S-VED, Chapbook, and HORAE datasets}
    \label{table:mAP_1000}
\end{table}

The results presented above are, despite their limitations, very promising, especially for researchers aiming to collect large image datasets from archival material at scale. It is clear that such models soon hit their limits when it comes to differentiating between image classes that might be of interest for historical research (e.g., Initials, Content Illustrations, and Decorations). However, the possibility of quickly collecting thousands of images from historical documents using descriptive language remains enticing, and will increase the efficiency of data collection, which is often a major barrier to applying ML algorithms in the frame of historical research. In this case, the scholars could invest human power on fine-tuning the retrieved data and creating well-curated sub-classes, which can then power the training of an in-domain model. \par

\subsection{A note on the environment}\label{text:env}
As we are living in a climate-critical era, it is imperative that we take environmentally conscious choices when dealing with computational data at scale. In this case, we acknowledge that the use of both GroundingDINO and SAM comes at a high computational cost. Although these models have zero-shot capabilities, which means we do not need to spend energy on training them, a single inference across this pipeline takes ca. 40 times longer (on CPU) than a single inference using models such as YOLOv8. Thus, we highly recommend using such a pipeline for preliminary data collection followed by training a specific-domain model that can then perform inferences at scale.

\section{Conclusion}\label{text:FutureDev}
In this paper, we explored the fast emerging field of multi-modal models and investigated its suitability for the digital humanist. The results of our investigation using the proposed pipeline show great potential from a technical aspect. We believe that this potential will lead to the generation of larger humanities datasets in the near future, but also to a larger interest and engagement from humanities scholars in computational approaches. This, we believe, is largely due to the linguistic interaction between the scholars and the machine, which is becoming one of the most human-computer interaction modes. This paper builds on the ``multimodal turn in the Digital Humanities'' \cite{Smits2023}. This language interaction also forces us, as digital humanists, to reconsider object and class definitions, and reformulate them in a more computer-suited linguistic approach, which can often be very challenging, and often lead to new definitions and hypotheses. \par

The work on this pipeline is part of an ongoing infrastructure project at the Max Planck Institute for the History of Science that aims to collect large amounts of visual content from heterogeneous historical documents. We used the pre-trained GroundingDINO as our object extractor in this paper; however, in the medium term we also plan to fine-tune this model on humanities-specific datasets in order to allow specific linguistic prompts to match the desired image region. In the long run, we plan to slowly build an application with a simple GUI around this pipeline to allow humanists with minimal computer science knowledge to extract such information from historical documents.

\section*{Funding}
This work was supported by the German Ministry for Education and Research as BIFOLD -- Berlin Institute for the Foundations of Learning and Data (grant 01IS18037A) and the Max Plank Institute for the History of Science.

\section*{Code availability}
The code for this pipeline is available here: \url{https://github.com/hassanhajj910/prompt-me-a-dataset}.

\section*{Acknowledgments}
We would like the thank Lindy Divarci and Luis Melendrez Zehfuss for the English proofreading. 
%
%
%
\bibliographystyle{splncs04}
\bibliography{bibliography}
%




\end{document}